\pdfoutput=1

\documentclass[11pt]{article}

\usepackage[]{emnlp2021}

\usepackage{times}
\usepackage{latexsym}

\usepackage[T1]{fontenc}

\usepackage[utf8]{inputenc}

\usepackage{microtype}
\usepackage{graphicx}

\makeatletter
\newcommand*\bigcdot{\mathpalette\bigcdot@{.8}}
\newcommand*\bigcdot@[2]{\mathbin{\vcenter{\hbox{\scalebox{#2}{$\m@th#1\bullet$}}}}}
\makeatother

\title{\includegraphics[scale=0.08]{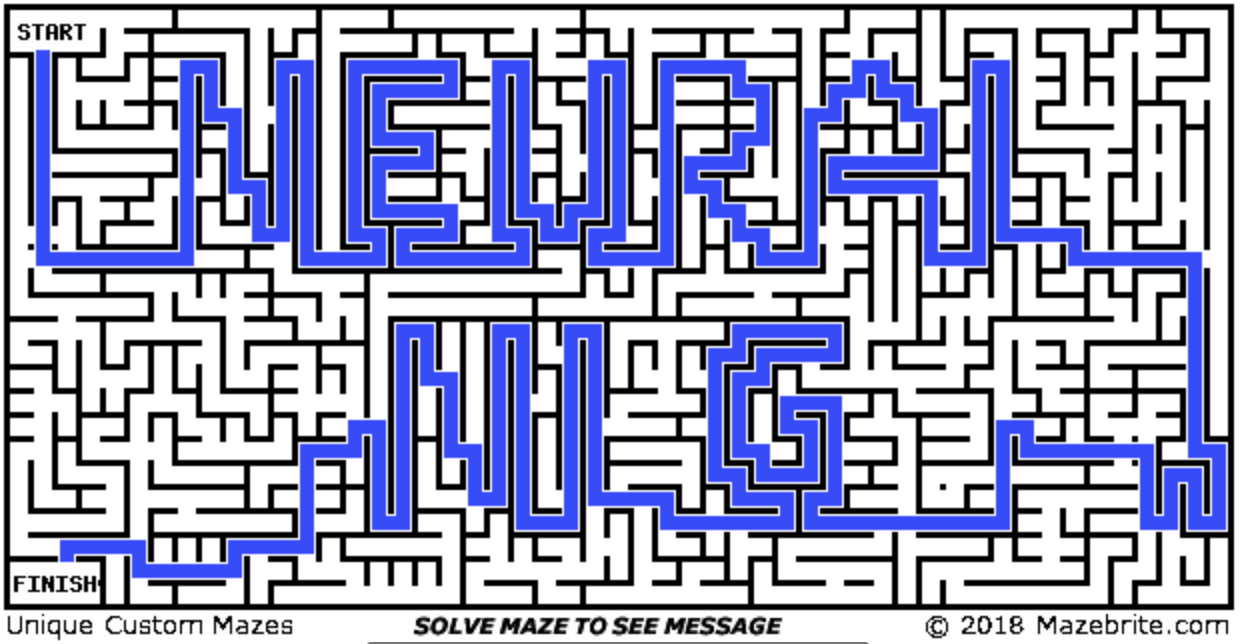} Positioning yourself in the maze of Neural Text Generation: \\A Task-Agnostic Survey}

\author{
Khyathi Raghavi Chandu  \qquad  Alan W  Black\\
\\
Language Technologies Institute, Carnegie Mellon University \\
\tt{ \{kchandu, awb\}@cs.cmu.edu }
}

\begin{document}
\maketitle
\begin{abstract}
Neural text generation metamorphosed into several critical natural language applications ranging from text completion to free form narrative generation. 
In order to progress research in text generation, it is critical to absorb the existing research works and position ourselves in this massively growing field.
Specifically, this paper surveys the fundamental components of modeling approaches relaying task agnostic impacts across various generation tasks such as storytelling, summarization, translation etc., 
In this context, we present an abstraction of the imperative techniques with respect to learning paradigms, pretraining, modeling approaches, decoding and the key challenges outstanding in the field in each of them. 
Thereby, we deliver a one-stop destination for researchers in the field to facilitate a perspective on where to situate their work and how it impacts other closely related generation tasks.

\end{abstract}

\section{Introduction}

The surge in neural methods for language processing deliberates us to cautiously foresee the upcoming challenges. One of the fields witnessing a steep growth here is text generation, which is the task of producing written or spoken narrative from structured or unstructured data. 
This field navigated through a variety of techniques and challenges from using template based systems, modeling discourse structures, statistical methods to more recent autoregressive deep nets, transformers etc.,
With this rapid transformation, it is critical to retrospect and position ourselves to foresee the upcoming task-agnostic challenges to impact the entire field. 
\textit{The primary goal of this paper is to assist the readers to position their work in this vast maze of text generation to identify new challenges and secondarily present a compact survey of the field in the context of task-agnostic challenges.}

\begin{figure}[t!]
\centering
\includegraphics[trim={2cm 3.2cm 1cm 2.0cm}, width=0.99\linewidth]{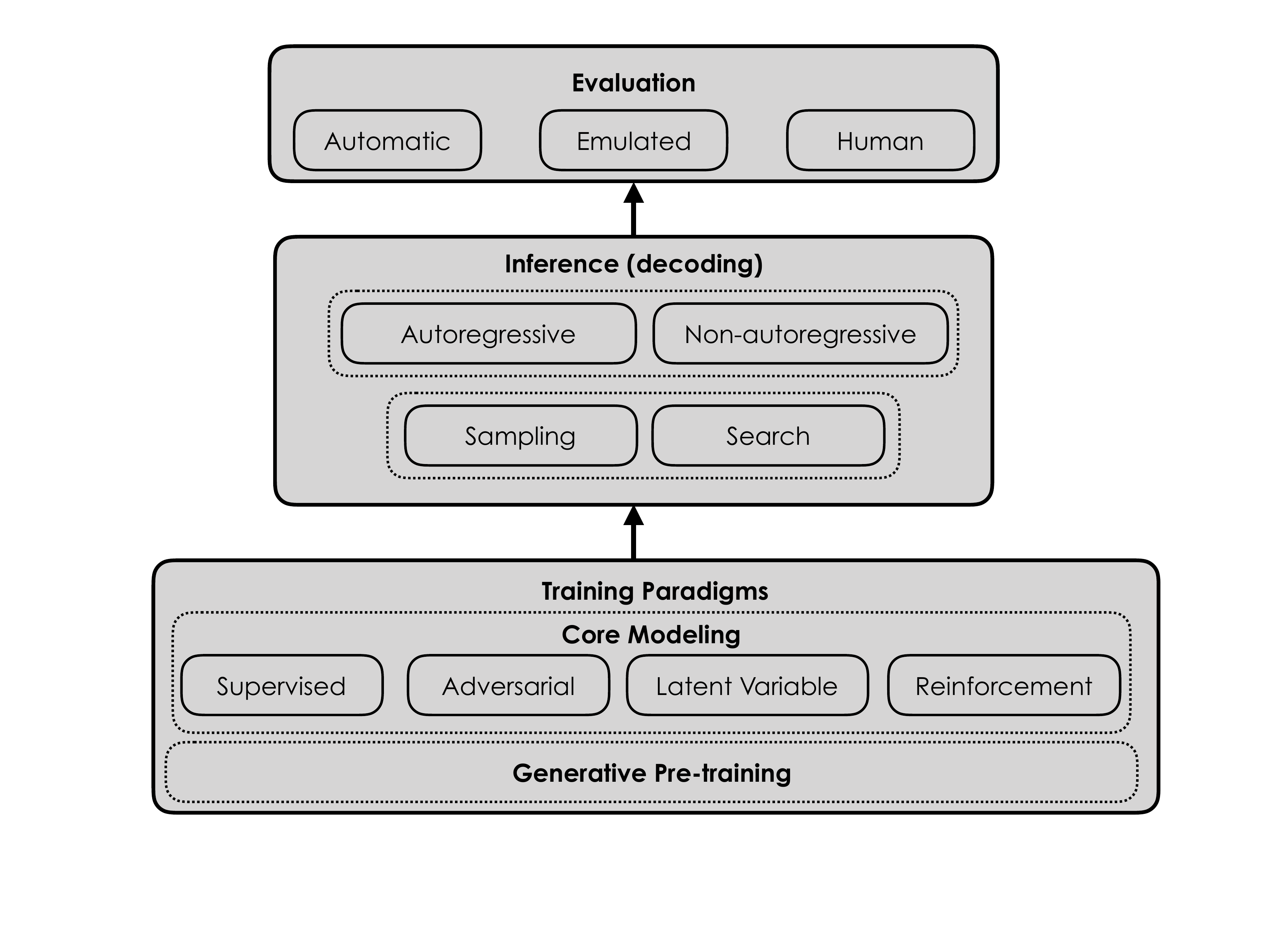}
\caption{Outlining the components of neural text generation discussed through the paper.}
\label{fig:topology}
\end{figure}

Before diving into the tasks-agnostic techniques, table \ref{tab:paradigms} presents the three main paradigms of tasks in generating text based on the schema of input and output. 
These categories are presented for the sake of completeness of the topic at high level but we do not go into their details in this paper.
These several tasks deserve undivided attention and accordingly they have been heavily surveyed in the recent past. For instance, independent and exclusive surveys are periodically conducted on summarization \cite{lin2019abstractive, allahyari2017text, nenkova2012survey, tas2007survey}, knowledge to text generation \cite{DBLP:conf/inlg/GardentSNP17, DBLP:conf/naacl/Koncel-Kedziorski19}, machine translation \cite{chu2018survey, dabre2019survey, chand2016empirical, slocum1985survey}, dialog response generation \cite{liu2016not, montenegro2019survey, ramesh2017survey, DBLP:journals/sigkdd/ChenLYT17}, storytelling, narrative generation \cite{DBLP:journals/information/TongRBWLWLWQLM18, DBLP:journals/tciaig/TogeliusYSB11}, image captioning \cite{DBLP:journals/corr/abs-1810-04020} etc., to dig deeper into task specific approaches that are foundational as well as in the bleeding edge of research. In addition, there have been several studies conducted on surveying text generation. \citet{DBLP:journals/cai/PereraN17} present a detailed overview of information theory based approaches. \citet{iqbal2020survey} primarily focus on core modeling approaches, 
\citet{DBLP:journals/jair/GattK18} elaborated on tasks such as captioning, style trasfer etc., with a primary focus on data-to-text tasks. Controllability aspect is explored by \citet{prabhumoye2020exploring}.  \citet{DBLP:journals/corr/abs-1803-07133} perform an empirical study on the core more modeling approaches only. While they are extremely necessary, the focus on techniques that are beneficial to other related tasks are often overlooked. This paper focuses on these task agnostic components to improve the ensemble of tasks in neural text generation. 

Figure \ref{fig:topology} presents the components that are important to study in neural text generation which are elaborated in this paper. Throughout the paper, we identify and highlight (in italics) the challenges in the field in the context of existing work.




\section{Training Paradigms}
\label{sec:model}

\subsection{Generative Pre-training}

\begin{table}[t]
\small
\centering
\resizebox{0.5\textwidth}{!}{%
\begin{tabular}{l|l|l|l}
\hline \hline
\bf Generation & \bf  & \bf  & \bf  \\
\bf Paradigm & \bf Task & \bf  Input & \bf  Output\\
\hline
Text-to-Text & Dialog & Conversation History & Next Response \\
&Machine Translation&	Source Language&	Target Language\\
&Style Transfer&	Style 1 Text&	Style 2 Text\\
&Summarization&	Single/Multiple Documents &	Summary \\
\hline
Data-to-Text & Image Captioning	& Image &	Descriptive Text\\
&Visual Storytelling & Images & Descriptive Text \\
&Speech Recognition &	Audio &	Text\\
&Table to Text&	Table&	Text\\
&Knowledge Bases to Text&	Knowledge Bases&	Text\\
\hline
None-to-Text &Language Modeling&	Null&	Sequence of Text \\
\hline
\hline
\end{tabular}
}
\caption{Paradigms of Tasks in Text Generation (not detailed in this paper). { \small Note: To be compact, we include `Knowledge-to-text' paradigm within `Data-to-text'.}}
\label{tab:paradigms}
\end{table}

Recent couple of years have seen a major surge in interest for pre-training techniques. 
UniLM (UNIfied pre-trained Language Model, \cite{dong2019unified}) is proposed as a pre-training mechanism for both natural language understanding and natural language generation tasks. Fundamentally, the previously widely used ELMO \cite{peters2018deep} constitutes a language model that is left to right and right to left. While GPT \cite{radfordimproving} has an autoregressive left to right language model, BERT \cite{DBLP:conf/naacl/DevlinCLT19} has a bidirectional language model. UniLM is optimized jointly for all of the above objectives along with an additional new seq2seq LM which is bidirectional encoding followed by unidirectional decoding. Depending on the use case, UniLM can be adopted to use Unidirectional LM,
Bidirectional LM 
and Seq2seq LM 
With a similar goal in mind, MASS \cite{song2019mass} modified masking patterns in input to achieve this. BERT and XLNet \cite{yang2019xlnet} pre-train an encoder and GPT pretrains a decoder. This is a framework introduced to pretrain encoder-attention-decoder together. Encoder masks a sequence of length k and the decoder predicts the same sequence of length k and every other token is masked. 
While the idea of jointly training the encoder-attention-decoder remains the same as in UniLM, the interesting contribution here is the way masking is utilized to bring out the following advantages. \textit{(i)} The tokens masked in decoder are the tokens that are not masked in encoder. 
\textit{(ii)} Encoder supports decoder by extracting useful information from the masked fragments improving the NLU capabilities.
\textit{(iii)} Since a sequence of length k is decoded consecutively, NLG capability is improved as well. 
Note that when k is 1, the model is closer to BERT which is biased to an encoder and when k is the length of sentence, the model is closer to GPT which is biased to decoder. Similar to UniLM, BART \cite{lewis2019bart} has a bidirectional encoder and an autoregressive decoder. The underlying model is standard transformer \cite{DBLP:conf/nips/VaswaniSPUJGKP17} based neural MT framework. The main difference of BART from MASS is that the tokens masked here are not necessarily consecutive. The main idea and is to corrupt text with arbitrary noise such as token masking, token deletion, token infilling, sentence permutation and document rotation and reconstruct original text. 
Following this, \citet{DBLP:journals/corr/abs-1910-10683} proposed T5 as a unifying framework that ties all NLP problems as text generation tasks with a text-in and text-out paradigm. Recently, \citet{DBLP:conf/iclr/DathathriMLHFMY20} introduced plug and play language models capable of efficiently training fewer parameters to control a huge underlying pretrained model. Finetuning these vast models for generative tasks has been studied in style transformers \cite{DBLP:conf/emnlp/SudhakarUM19} and conversational agents \cite{DBLP:conf/iclr/DinanRSFAW19}.

\noindent \textbf{Challenges: } 
This new era of very powerful language models opened up a whole new set of challenges. \textit{How can they be effectively used as knowledge sources \cite{DBLP:journals/corr/abs-2005-11401} ?}
\textit{How to mitigate the inherent societal biases from models learnt at this scale \cite{DBLP:conf/emnlp/ShwartzRT20} ?} 
\textit{How to learn social norms from vast amounts of pretrained models \cite{DBLP:conf/emnlp/ForbesHSSC20} ?}
\textit{How to ensure coherence in long form generation specific to a domain \cite{DBLP:journals/corr/abs-2006-15720} ?}


\subsection{Core Modeling}
The base architecture constitutes of an encoder-decoder with an optional attention mechanism.

\noindent \textbf{Supervised Learning: }
Most generation approaches in this setting use maximum likelihood objective for training sequence generation with a sequential multi-label cross entropy. 
However, there is an inherent inconsistency in exposure to ground truth text between training and inference stages when using teacher forcing during training. 
leading to exposure bias \cite{DBLP:journals/corr/RanzatoCAZ15}. 
This problem becomes severe with the increasing length of the output. A solution to address this issue is scheduled sampling \cite{DBLP:conf/nips/BengioVJS15} which mixes teacher forced embeddings and model predictions from previous time step. \textit{This problem of text degeneration still remains a pressing issue}, which is revisited in \S \ref{sec:key_challenges}.

\noindent \textbf{Reinforcement Learning: }
The main issue with the supervised learning approach for text generation is the mismatch between maximum likelihood objective and metrics for text quality. Reinforcement learning addresses this mismatch by directly optimizing end metrics which could be non-differentiable. Typically, policy gradient algorithms are used to optimize for BLEU score directly via Reinforce.
However, \textit{computing these metrics before every update is not computationally efficient to incorporate in the training procedure}. Another problem is the \textit{inherent inefficiency of the metric itself used for reward} i.e BLEU is not the best measure to evaluate text quality. Sometimes these rewards are learnt adversarially. In practice, usually, the policy network is usually pre-trained with maximum likelihood objective before optimizing for BLEU score.

\noindent \textbf{Latent Variable Modeling: }
These models took a pretty steep curve from variational seq2seq models \cite{DBLP:conf/conll/BowmanVVDJB16} to conditional VAEs, \cite{DBLP:conf/acl/ShenSLLNZAL17}. These latent variable models have been explored to generate controllable text generation based on topic \cite{DBLP:conf/aaai/WangTHQZL19}, structure \cite{DBLP:conf/acl/ChenTWG19}, persona \cite{DBLP:conf/acl/ZhaoZE17} \cite{DBLP:conf/acl/WuLWCWFHW20} etc., 

\noindent \textbf{Adversarial Learning: }
The third paradigm is adversarial learning comprising of competing objectives. The mismatch in training and inference stages is addressed using Professor Forcing \cite{lamb2016professor} with adversarial domain adaptation to bring the behavior of the training and sampling close to each other. 
Generative Adversarial Networks (GAN) also gained popularity with respect to this in the recent times. The core idea is that the gradient of the discriminator guides how to alter the generated data and by what margin in order to make it more realistic. 
There are several variants adopted to address specific problems such as SeqGAN to assess partially generated sequence \cite{DBLP:conf/aaai/YuZWY17}, MaskGAN to improve sample quality using text filling \cite{DBLP:conf/iclr/FedusGD18} and LeakGAN to model long term dependencies by leaking discriminator information to generator \cite{DBLP:conf/aaai/GuoLCZYW18}. The three main challenges researched in this area are:

\noindent $\bigcdot$ \textit{Discrete Sampling: } The sampling step selecting argmax in language is non-differentiable. This is addressed by replacing it with a continuous approximation by adding Gumbel noise which is negative log of negative log of a sample from uniform distribution, also known as Gumbel Softmax.

\noindent $\bigcdot$ \textit{Mode Collapse: } GANs typically face the issue of sampling from specific tokens to cheat discriminator, known as mode collapse. In this way, only a subspace of target distribution is learnt by the generator. DP-GAN addresses this using an explicit diversity promoting reward \cite{DBLP:journals/corr/abs-1802-01345}.

\noindent $\bigcdot$ \textit{Power dynamics between Generator and Discriminator during training:} Another problem arises when the discriminator is trained faster than the generator and overpowers it. This phenomenon in training is observed frequently, and the gradient from discriminator vanishes leading to no real update to generator.

\subsection{Decoding Strategies}

The natural next step after pre-training and training is decoding. The distinguishing characteristic of generation is the absence of one to one correspondence between time steps of input and the output, thereby introducing a crucial component which is decoding. Primarily, they can be categorized as (i) autoregressive and (ii) non-autoregressive. 

\noindent \textbf{Autoregressive decoding: } Traditional models with this strategy correspond well to the true distributions of words. This mainly comes from respecting the conditional dependence property from left to right. The autoregressive techniques can be further viewed as sampling and search techniques. One of the issues of this strategy is \textit{throttling transformer based models that fall short in replicating their training advantages as training can be non-sequential} and inference holds to be sequential with autoregressive decoding.

\noindent \textbf{Non-autoregressive decoding: } This line of work primarily addresses two problems that are associated with autoregressive decoding. First, by definition, there is a conditional independence property that holds. This leads to the \textit{multimodality problem, where each time step considers different variants with respect to the entire sequence} and these conditions compete with each other. Second, the main advantage is the reduction in latency during real time generation. \citet{DBLP:conf/aaai/GuoTXQCL20} addressed this problem in the context of neural machine translation using transformers by copying each of the source inputs to the decoder either uniformly or repeatedly based on their fertility counts. 
These fertilities are predicted using a dedicated neural network to reduce the unsupervised problem to a supervised one and thereby enabling it to be used as a latent variable. This invariable replications based on fertilities may lead to \textit{repetition or duplication of words}. Closely followed by this, \citet{DBLP:conf/icml/OordLBSVKDLCSCG18} took a different approach by introducing probability density distillation by modifying a convolutional neural network using a pre-trained teacher network to score a student network attempting to minimize the KL divergence between itself and the teacher network. Both these works set the trend of using latent variables to capture the interdependence between different time steps in the decoder. Following this work,  \citet{DBLP:conf/emnlp/LeeMC18} use iterative refinement by denoising the latent variables at each of the refinement steps. This idea of iterative decoding inspired way to more avenues by combining the benefits of cloze style mask prediction objectives from Bert \cite{DBLP:conf/naacl/DevlinCLT19}.  Some of them include insertion based techniques \cite{DBLP:journals/tacl/GuLC19}, repeated masking and regenerating \cite{DBLP:conf/emnlp/GhazvininejadLL19} and  providing model predictions to the input \cite{ghazvininejad2020semi}.
\citet{DBLP:conf/aaai/WangTHQZL19} proposed an alternative approach to address repetition (observed in \cite{DBLP:conf/aaai/GuoTXQCL20}) and completeness using regularization terms for each. Repetition is handled by regularizing similarity between consecutive words. Completeness is addressed by enabling reconstruction of source sentence from hidden states of the decoder.
Concurrently, \citet{DBLP:conf/aaai/GuoTHQXL19} also address these issues by improving the inputs to decoder using additional phrase table information and sentence level alignment.

\noindent \textbf{Sampling and Search Techniques:}

\noindent \textbf{1. Random Sampling: } The words are sampled randomly based on the probability from the entire distribution without pruning any of the mass.
    
\noindent \textbf{2. Greedy Decoding: } This technique simply boils down to selecting argmax of the probability distribution. As you keep selecting argmax everywhere, the problem is that it limits the diversity of generation. Temperature scaling helps to adjust the spectrum between flat and peaky distributions to generate more diverse or safe responses respectively.
This is alleviated by the next techniques and beam search. This is also worked out for discrete settings using gumbel-greedy decoding \cite{gu2018neural}. Variants of this were also studied by \citet{zarriess2018decoding}

\noindent \textbf{3. Beam Search: } Beam search introduces a course correction mechanism in approximation of the argmax by selecting a beam size number of beams at each time step. 
It has been relatively well studied in task agnostic objectives \cite{wang2014beam} for instance, including social media text \cite{wang2013beam}, error correction \cite{dahlmeier2012beam}. \textit{Small beam sizes may lead to ungrammatical sentences}, they get more grammatical with increasing beam size. Similarly small beam sizes may be less relevant with respect to content but get more \textit{generic with increasing beam size}. Prominent variants within beam search are:

    \noindent (a) Noisy Parallel Approximate Decoding:  This method \cite{DBLP:journals/corr/Cho16} introduces some noise in each hidden state to non-deterministically make it slightly deviate from argmax.
    
    \noindent (b) Beam Blocking:  Repetition is one of the problems we see in NLG and this technique \cite{DBLP:conf/iclr/PaulusXS18} combats this problem by blocking the repeated n-grams. It essentially adjusts the probability of any repeated n-gram to 0.
    
    \noindent (c) Iterative Beam Search: In order to search a more diverse search space, another technique \cite{kulikov2019importance} was introduced to iteratively perform beam search several times. And for each current time step, we avoid all of the partial hypotheses encountered until that time step in the previous iterations based on soft or hard decisions on how to include or exclude these beams.
    
    \noindent (d) Diverse Beam Search: One problem with beam search is that most times the decoded sequence still tends to come from a few highly significant beams thereby suppressing diversity. The moderation by \cite{vijayakumar2016diverse} adds a diversity penalty computed (for example using hamming distance) between the current hypothesis and the hypotheses in the groups to readjust the scores for predicting the next word.
    
    \noindent (e) Clustered Beam Search:  The goal is prune unnecessary beams. At each time step, \citet{tam2020cluster} get the top candidates and embed them by using averaged Glove representations which are clustered using k-means to pick from each cluster.
    
\noindent (f) Clustering Post Decoding:  
This technique \cite{kriz2019complexity} clusters after decoding as opposed to modifying the decoding step. Sentence representations from any of the diversity promoting beam search variants are obtained. These are then clustered and the sentence with high log likelihood is selected from the cluster.
    

\noindent \textbf{4. Top-k sampling: } This technique by \citet{fan2018hierarchical} samples from the k most probable candidates from the output distribution. This means that we are confining the model to select from a truncated probability mass.
\noindent If $k$ is the size of vocabulary, then it is random sampling and if $k$ is 1 then it is greedy decoding. High valued k results in dicey words but are non-monotonous and low valued k results in safe outputs which are monotonous. The problem however is that k is limited to the same value in all scenarios.

\noindent \textbf{5. Top-p sampling: } The aforementioned problem of a fixed value of $k$ is addressed by top-p sampling. This is also known as nucleus sampling \cite{DBLP:conf/iclr/HoltzmanBDFC20}, which instead of getting rid of the unspecified probability mass in top-k sampling, importance is shifted to the amount of probability mass preserved. This addresses scenarios where there could be broader set of reasonable options and sometimes a narrow set of options. It is achieved by selecting a dynamic $k$ number of words from a cumulative probability distribution until a threshold probability value is attained.

\section{Key Challenges}
\label{sec:key_challenges}

For each of the challenges, this section provides a list of solutions. The pitfalls of these solutions are also described there by encouraging research to address these key challenges.

\paragraph{1. Content Selection: }
Certain tasks demand copying over the details in the input such as rare proper nouns for instance in news articles etc., This is especially needed in tasks like summarization which can demand a combination of extractive and abstractive techniques.

\noindent $\bigcdot$ \textbf{Copy Mechanism: } Copy mechanism can take various forms such as pointing to unknown words \cite{gulcehre2016pointing} based on attention \cite{see2017get} or a joint or a conditional copy mechanism \cite{gu2016incorporating, puduppully2019data}. It maybe based on attention that copies segments from input into the output. \textit{The challenge in this technique is to make sure that this combination of being extractive and abstractive does not boil down to a purely extractive system.}

\noindent $\bigcdot$ \textbf{Attention mechanism: } This is actively used for subselecting content in tasks such as summarization \cite{DBLP:conf/naacl/ChopraAR16}. Recent work has demonstrated instances of \textit{attention not explaining the output} \cite{jain2019attention, DBLP:journals/corr/abs-2004-05569}.

\noindent $\bigcdot$ \textbf{Hierarchical Modeling: } This technique maintains a global account of the content. This is often modeled using hierarchical techniques or dual stage models \cite{martin2018event, xu2018skeleton, gehrmann2018bottom} where the first stage pre-selects relevant keywords for generation in the following stage. 
\textit{Such models possibly take a hit on fluency while connecting dots between selected content and generation.} This means that Rouge-1 can be good because the right words are extracted but Rouge-2 may decrease as it affects the fluency.

\noindent $\bigcdot$ \textbf{Memory Modules: } 
\cite{DBLP:conf/aaai/ZhouHZZL18} and \cite{DBLP:conf/naacl/ClarkJS18} explored memory modules in inducing emotion and entity representations from external memory respectively.
\textit{An outstanding challenge still remains in exploring the best ways to encode this external memory.}

\paragraph{2. Repetition:}
\citet{DBLP:conf/iclr/HoltzmanBDFC20} demonstrate that the objective of maximum likelihood renders high log likelihood for the words that have been generated, leading to repetition. This problem amplifies with increasing sequence length and transformer based models.

\noindent $\bigcdot$ \textbf{Beam blocking: }Blocking beams containing previously generated n-grams from subsequent generation combats repetition and encourages diversity 
\cite{klein2017opennmt, DBLP:conf/iclr/PaulusXS18} etc., 
\textit{Selecting the number of beams is often a problem since it is natural for a function word to repeat more often.}
\citet{massarelli2019decoding} extensively studied the variants of n-gram blocking by applying delays in beam search.

\noindent $\bigcdot$ \textbf{Unlikelihood objective: } \citet{DBLP:conf/iclr/WelleckKRDCW20} argue that there is a fundamental flaw in the objective of maximum likelihood. 
The main idea is to decrease the probability of unlikely or negative candidates. The negative candidates are selected from the previous contexts either at token or at sequence levels which are essentially n-grams. 
\textit{Selecting negative contexts is tricky and needs to be beyond selection of simple n-gram sequences that occurred previously. }

\noindent $\bigcdot$ \textbf{Coverage penalty: } This discourages the attention mechanism to attend the same word repeatedly \cite{see2017get} by assigning coverage penalty of the attention probability mass on that source time step for each decoded time step.

\paragraph{3. Coherence:}
This is a critical property of text to factor in for multi-sentence or long form generation, that not only takes into account the appropriate content but also the structure of the narration.

\noindent $\bigcdot$ \textbf{Static and Dynamic Planning: } This addresses coherence in terms of layout or structural organization of the text \cite{yao2019plan}. A schema of static or dynamic plans are used to form an abstract flow of the text from which the actual text is realized. 
\textit{However, underlying language models are capable of taking over, leading to hallucinations and thereby compromising the fidelity of text.}

\paragraph{4. Length of Decoding: }
One factor that distinguishes generation from rest of the seq2seq family of tasks is the variability in the length of the generated output. The main problem here is that as the length of the sequence increases, the sum of the log probability scores decrease. This means that \textit{models prefer shorter hypotheses}.
Some solutions to combat this problem are the following. 

\noindent $\bigcdot$ \textbf{Length Normalization or Penalty:} The generated output is scored by normalizing or dividing with length. \cite{wu2016google} explore a different variation of the normalization constant. 
    
\noindent $\bigcdot$ \textbf{Probability boosting: } This technique multiplies the probability with a fixed constant at every time step. This alleviates the diminishing score problem. 

\noindent $\bigcdot$ \textbf{ Length based bias: } Incorporate  bias in the model based on empirical relations on lengths in source and target sentences in the training data.

\paragraph{5. Sequence level scoring: }
Instead of modifying the decoding, this strategy perform sequence level scoring from multiple texts deccoded.

\noindent $\bigcdot$ \textbf{Reranking: } Another mechanism is to sample several full sequences and rerank them based on generic scores such as perplexity or BLEU or task specific requirements varying from factual correctness \cite{DBLP:conf/emnlp/GoyalD20}, coherence \cite{DBLP:journals/corr/abs-2010-12884}, style \cite{DBLP:conf/acl/ChoiBGHBF18} etc., \textit{The properties of text in the end goal are decoupled from the generation process} (a soft conditioning of reranking on generation helps improve the generation as well).

\paragraph{5. Optimization Objective: }
Similar to the observation earlier in Section \ref{sec:model}, there is an inherent mismatch in the between the objective function which is maximum likelihood and the end metrics which are BLEU, Rouge etc;

\noindent $\bigcdot$ \textbf{Reinforcement Learning: } A common solution for this problem is using reinforcement learning to optimize end metrics such as Rouge. Often, a combination of MLE and RL objectives are used \cite{DBLP:conf/aaai/HuCGLGN20, wang2018no} to optimize BLEU \cite{DBLP:journals/corr/WuSCLNMKCGMKSJL16}, ROUGE \cite{DBLP:conf/iclr/PaulusXS18}, CIDEr \cite{DBLP:conf/cvpr/RennieMMRG17}, SPIDEr \cite{DBLP:conf/iccv/LiuZYG017}. \textit{An existing open challenge is to understand how to make the models robust by making them learn the task rather than gaming for the reward.} The rewards can also be learnt adversarially during the training \cite{DBLP:conf/emnlp/LiMSJRJ17}.
\textit{However, this is still a problem since these end metrics do not directly correlate to human judgements. }


\noindent $\bigcdot$ \textbf{Factorizing Softmax: } \cite{DBLP:conf/emnlp/ChoiHPL20} recently proposed a method to factorize softmax by learning to predict both the frequency class and the token itself during training by factorizing the probabilities. \textit{This model is observed to repeat the same rare token across several sentences.}

\noindent $\bigcdot$ \textbf{Maximum Mutual Information: } The idea is to incorporate pairwise information of source and target instead of only one direction which is usually target given source \cite{li2016diversity}. The target probability is subtracted from target given source probability to diminish the probability of generic sentences. \textit{The model optimized with MMI can sometimes generate ungrammatical sentences.}

\noindent $\bigcdot$ \textbf{Distinguishability: } Hallucinations in abstractive generation are unwanted byproducts of optimizing log loss. To combat this, several researchers explored optimizing for minimized distinguishability with human generated text \cite{DBLP:conf/naacl/HashimotoZL19, DBLP:journals/corr/TheisOB15}. Following similar path,   \citet{DBLP:journals/corr/abs-2004-14589} proposed truncating loss to get rid of unwanted samples. 

\noindent \textbf{5. Speed:}
Practical applications call for a crucial research direction of generating text in real time in addition to chasing the state of the art results. Model compression plays a crucial part in demonstrating an increase in the speed of generation. \citet{cheng2017survey} exhaustively surveyed the different techniques to perform model compression. While there are techniques in the hardware side, there are certain modeling approaches that can handle this problem as well \cite{gonzalvo2016recent}. Most of this work is studied in the context of real time interpretation of speech \cite{fugen2007simultaneous, yarmohammadi2013incremental, grissom2014don}. Recently, \citet{DBLP:journals/corr/abs-2006-01112} proposed a cascaded decoding approach introducing Markov Transformers to demonstrating high speed and accuracy. 

\noindent $\bigcdot$ \textit{Quantization: } Quantizing \cite{DBLP:journals/corr/abs-1805-11063, gray1984vector} the weights i.e sharing the same weight when they belong to a bin proved helpful in improving the speed. This also facilitates the computations of gradients only once per bin.

\noindent $\bigcdot$ \textit{Distillation: } It can be performed with a teacher and a smaller student network that tries to replicate the performance of the teacher with fewer parameters \cite{DBLP:journals/corr/abs-1911-03829, DBLP:conf/acl/ChenGCLL20}.

\noindent $\bigcdot$ \textit{Pruning: } This technique thresholds and prunes all the connections that have weights lesser than the predetermined threshold and then we can retrain the network in order to adjust the weights of the remaining connections.

\noindent $\bigcdot$ \textit{Real time: } \citet{DBLP:conf/eacl/NeubigCGL17} trained an agent that learns to decide between the actions of reading by discarding a candidate or writing by accepting a candidate. The policy network is optimized with a combination of quality evaluated with BLEU and delay evaluated by number of consecutive words in reading stage which increases wait time.

\noindent $\bigcdot$ \textit{Caching: } Another trick is to cache some of the previous computations to avoid repetition.

\section{Evaluation}
\label{sec:eval}
Similar to other generative modeling, text generation also faces crucial challenges in evaluation \cite{reiter2009investigation, DBLP:journals/coling/Reiter18}. \citet{DBLP:conf/inlg/LeeGMWK19} present some of the best practices of evaluating automatically generated text. The main hindrance to standardize or evaluate NLG like other standard tasks is that it is often a sub-component of other tasks. \citet{DBLP:journals/corr/abs-2006-14799} present a more comprehensive survey of evaluation metrics for text generation

\paragraph{Desiderata of Text: } It is crucial to define the factors contributing to the quality of good text. Some of the factors include relevant content, appropriate structure in terms of coherence and suitable surface forms. In addition, fluency, grammaticality, believability and novelty in some scenarios are crucial factors. 

\paragraph{Intrinsic and Extrinsic: }Evaluation in subjective scopes such as text generation can be performed intrinsically or extrinsically. Intrinsic evaluation is performed internally with respect to the generation itself and extrinsic evaluation is typically performed on the metric used to evaluate a downstream task in which this generation is used. The quality can also be judged using automatic metrics and human evaluation.

\paragraph{(a) Automatic Metrics:} 
These metrics can be classified into the following categories:

\noindent \textbf{ $\bigcdot$ Word overlap based metrics: } These are based on the extent of word overlap, which means that they capture replication of words. The problem with such measures is that they do not focus on semantics but rather just the surface form of words and alone. This includes precision for n-grams(BLEU \cite{DBLP:conf/acl/PapineniRWZ02}), self-BLEU \cite{DBLP:conf/sigir/ZhuLZGZWY18}, improved weighting for rare n-grams (NIST \cite{doddington2002automatic}), recall for n-grams (ROUGE \cite{lin2002manual}), F1 equivalent of n-grams (METEOR \cite{DBLP:conf/acl/BanerjeeL05}), tf-idf based cosine similarity for n-grams (CiDER \cite{DBLP:conf/cvpr/VedantamZP15}). In extension to this, we also have specific metrics to evaluate content selection by measuring summarization content units using PYRAMID \cite{DBLP:conf/naacl/NenkovaP04} and parsed scene graphs with objects and relations using SPICE \cite{DBLP:conf/eccv/AndersonFJG16}. \citet{DBLP:conf/wmt/StanojevicS14} proposed BEER to address this as a ranking problem with character n-grams along with words.

\noindent \textbf{ $\bigcdot$ Language Model based metrics: } This includes perplexity \cite{DBLP:journals/coling/BrownPPLM92}. Such metrics are good in commenting about the language model itself. It sort of gives the average number of choices each random variable has. However, it does not directly evaluate the generation itself, for instance a decrease in perplexity does not imply a decrease in the word error rate. \textit{The problem remains that this metric intrinsically conveys if the LM is good enough to select the right next word for that corpus but not the actual quality of the generated text.} The human likeness is also measured by training a model to discriminate between human and machine generated text such as an automatic turing test \cite{DBLP:conf/acl/LoweNSABP17, DBLP:conf/cvpr/CuiYVHB18, DBLP:conf/naacl/HashimotoZL19}.

\noindent \textbf{ $\bigcdot$ Embedding based metrics: } This has the advantage of being able to capture distributed semantics compared to word overlap metrics and language model based metrics. 
MEANT 2.0 \cite{DBLP:conf/wmt/Lo17} and YISI-1 \cite{DBLP:conf/wmt/LoSSLGL18} computes structural similarity with shallow semantic parses being definitely and discretionarily used respectively  along with word embeddings. Recently, contextulaized embeddings have been extensively used to capture this, such as BertScore \cite{DBLP:conf/iclr/ZhangKWWA20} and BLEURT \cite{DBLP:journals/corr/abs-2004-04696}. Metrics based on a combination of different embeddings are also proposed \cite{DBLP:conf/wmt/ShimanakaKK18, DBLP:conf/wmt/MaGWL17}. \textit{However the problem of not correlating to human judgements still persists.}

\paragraph{(b) Emulated Automatic Metrics: } These metrics check for the intended behavior in generation based on the specific sub-problem being addressed . 
For instance, diversity can be evaluated by computing corpus based distributions on number of distinct entities \cite{DBLP:conf/acl/FanLD19, DBLP:journals/corr/abs-1909-09699, DBLP:conf/naacl/ClarkJS18} and so on. Recent approaches worked on identifying factual inconsistencies with a QA model using QAGS \citet{wang2020asking}, answering cloze style questions using SummaQA \cite{DBLP:conf/emnlp/ScialomLPS19}, performance on a language understanding task using BLANC \cite{DBLP:journals/corr/abs-2002-09836}, adhering to pre-defined commonsense conditions \cite{DBLP:journals/corr/abs-2010-12834}. 

\paragraph{(c) Human Evaluation: } There are broadly two mechanisms in conducting subjective evaluations which is a challenging component of text generation. The first is preference testing and the second is scoring. Studies have shown that preference based testing is prone to high variance compared to absolute scoring. Here are some important points to keep in mind during conducting human evaluation. There are several problems with human evaluations. They are expensive, no universally agreed upon guidelines for setup, difficult to ensure quality control, varying scores based on scales (binary vs continuous), difficult to replicate, presenting the task in an unambiguous way. In order to measure more reliably, we need to collect multiple scores and compute inter-annotator agreement with  Cohen's, Krippendorff's coefficients etc.,
Having critically discussed human evaluation, this is still really the best we got. It is absolutely crucial to perform human evaluation in most tasks. So, the aforementioned problems need to be taken merely as cautions to develop 
rational and systematic testing conditions. Comparisons between automatic and human evaluation metrics \cite{DBLP:conf/eacl/BelzR06} are actively studied in order to bring human evaluations closer to automatic metrics.

\section{Conclusions }
\label{sec:conclusion}

The past decade witnessed text generation dribbling from niche scenarios into several mainstream NLP applications. This urges the need for a snapshot to retrospect the progress of varied text generation tasks in unison. 
Throughout the paper, we highlighted some of the existing challenges in each of the three main task-agnostic components of generation: training paradigms including generative pre-training, decoding strategies and evaluation. In addition, this paper also includes a dense account of key challenges along with some techniques to address them. 
We believe this paper manifests as a compact resource for major outstanding challenges in this vast maze of neural text generation that can assist researchers foraging to situate their work in a task-agnostic manner to accelerate the progress in future.
We believe understanding this space helps us foresee upcoming challenges in context by broad audiences, researchers and practitioners, in academia and industry.
Moving forward, we envision that there are some crucial directions to focus for impactful innovation in text generation. These include \textit{(i) generation in real time (ii) non-autoregressive decoding (iii) grounding with situated contexts in real and virtual environments and games (iv) consistency with personality and opinions especially for virtual agents (v) conditioning on multiple modalities together with text and data (vi) investigating better metrics to evaluate generation to correlate better with human judgements (vii)  creative text generation, such as jokes, sarcasm, metaphors etc., (viii) utilizing large scale pre-trained language models as knowledge sources, (ix) reducing societal biases and generate text ethically, (x) setting up benchmarks to evaluate generation agnostic and specific to individual tasks.} We believe this is the right time to extend advancements in any particular task to other tightly coupled tasks in order to revamp improvements in text generation as a holistic task in the future.

\bibliography{anthology,custom}
\bibliographystyle{acl_natbib}




\end{document}